\pdfoutput=1

\documentclass[11pt]{article}

\usepackage{acl}

\usepackage{times}
\usepackage{latexsym}

\usepackage[T1]{fontenc}

\usepackage[utf8]{inputenc}

\usepackage{microtype}

\usepackage{multirow}

\usepackage{booktabs}
\usepackage{lipsum}  
\usepackage{graphicx}
\usepackage{wrapfig}
\usepackage{colortbl}
\usepackage{array}
\usepackage{float}
\usepackage{capt-of}
\usepackage{url}
\usepackage{xurl}
\usepackage{amsmath}
\usepackage{amsfonts}
\usepackage{inconsolata}

\newcommand\blfootnote[1]{%
  \begingroup
  \renewcommand\thefootnote{}\footnote{#1}%
  \addtocounter{footnote}{-1}%
  \endgroup
}

\newcolumntype{C}[1]{>{\centering\let\newline\\\arraybackslash\hspace{0pt}}m{#1}}

\title{Sparse Distillation: Speeding Up Text Classification \\by Using Bigger Student Models}

\author{Qinyuan Ye$^{1\dagger}$  \quad Madian Khabsa$^{2}$ \quad Mike Lewis$^{2}$ \\ \textbf{Sinong Wang$^{2}$ \quad Xiang Ren$^{1}$ \quad Aaron Jaech$^{2}$}\\
$^{1}$University of Southern California \quad $^{2}$Meta AI\\
\texttt{\{qinyuany,xiangren\}@usc.edu} \\ \texttt{\{mkhabsa,mikelewis,sinongwang,ajaech\}@fb.com}
}

\begin{document}

\maketitle

\begin{abstract}


Distilling state-of-the-art transformer models into lightweight student models is an effective way to reduce computation cost at inference time. 
The student models are typically compact transformers with fewer parameters, while expensive operations such as self-attention persist. 
Therefore, the improved inference speed may still be unsatisfactory for real-time or high-volume use cases. 
In this paper, we aim to further push the limit of inference speed by distilling teacher models into bigger, sparser student models~--~bigger in that they scale up to billions of parameters; sparser in that most of the model parameters are n-gram embeddings. 
Our experiments on six single-sentence text classification tasks show that these student models retain 97\% of the RoBERTa-Large teacher performance on average, and meanwhile achieve up to 600x speed-up on both GPUs and CPUs at inference time. 
Further investigation reveals that our pipeline is also helpful for sentence-pair classification tasks, and in domain generalization settings.\blfootnote{\hspace{-0.1cm}$^\dagger$Work partially done while interning at Meta AI.}\footnote{Code available at \url{https://github.com/INK-USC/sparse-distillation}.}
\end{abstract}


\section{Introduction}

Large pre-trained Transformers \cite{devlin2018bert,liu2019roberta} are highly successful, but their large inference costs mean that people who host low-latency applications, or who are simply concerned with their cloud computing costs have looked for ways to reduce the costs.
Prior work mainly achieves this by leveraging knowledge distillation \cite{hinton2015distilling}, which allows for the capabilities of a large well-performing model known as the teacher to be transferred to a smaller student model. For example, DistillBERT \citep{sanh2019distilbert} is a smaller transformer model distilled from BERT \citep{devlin2018bert}, which reduces BERT's size by 40\% and becomes 60\% faster during inference. However, such speed-up may be still insufficient for high-volume or low-latency inference tasks. In this paper, we aim to further push the limit of inference speed, by introducing Sparse Distillation, a framework that distills the power of state-of-the-art transformer models into a shallow, sparsely-activated, and richly-parameterized student model. 

\begin{figure}[t]
    \centering
    \includegraphics[width=0.49\textwidth]{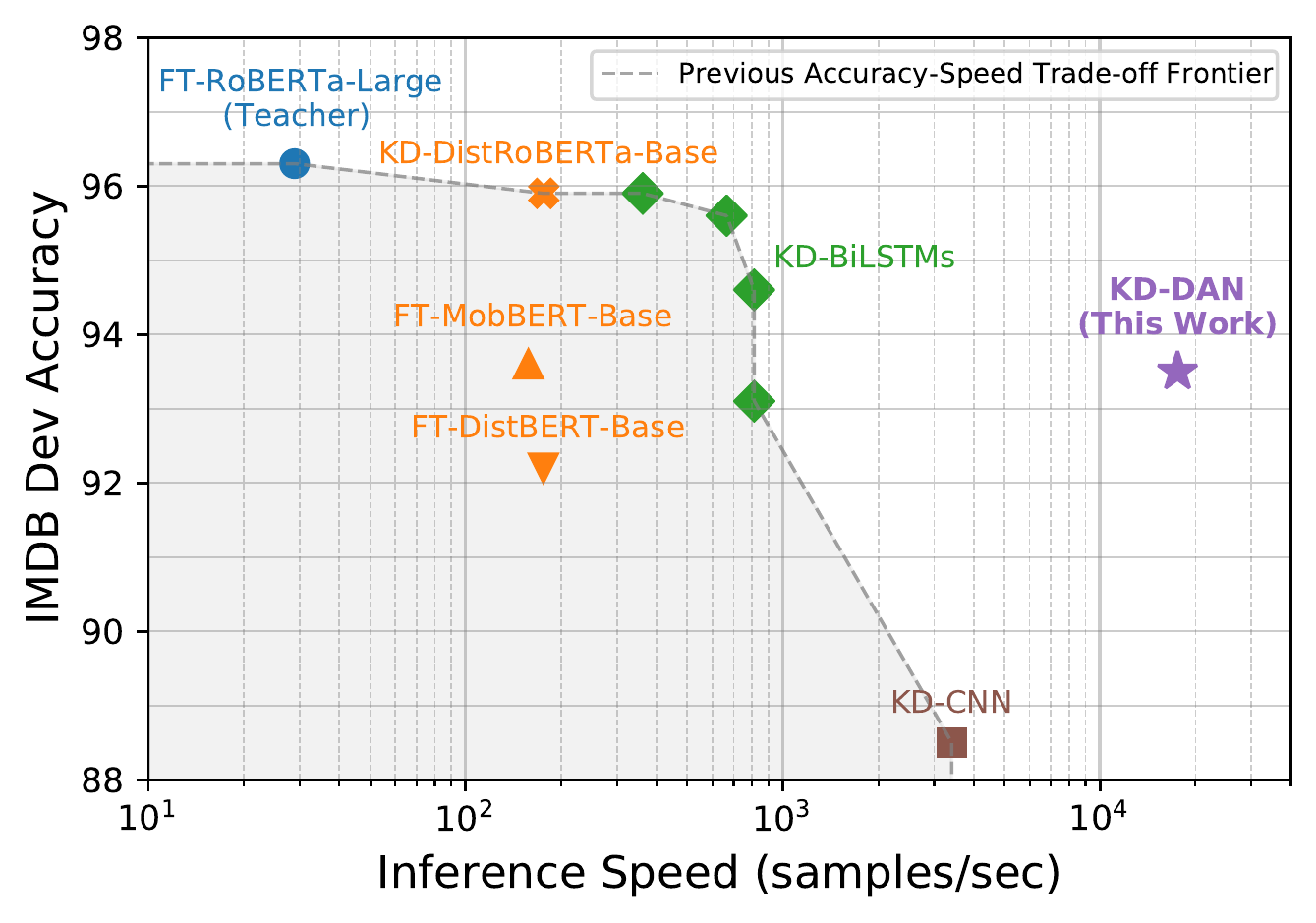}
    \vspace{-0.3cm}
    \caption{\textbf{Performance vs. Inference Speed.} With Deep Averaging Network (DAN; \citealt{iyyer2015deep}) and knowledge distillation, we obtain a student model with competitive performance on IMDB dataset, while being 607x faster than RoBERTa-Large, and 20x faster than bi-directional LSTMs at inference time.}
    \label{fig:speed}
\end{figure}

\begin{figure}[t]
    \centering
    \includegraphics[width=0.5\textwidth]{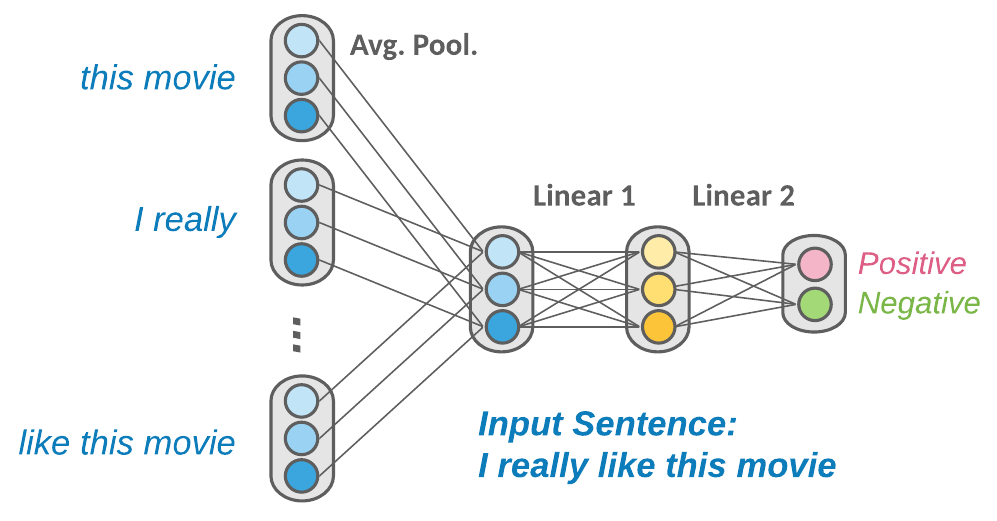}
    \caption{We primarily use a modified \textbf{Deep Averaging Network} (DAN; \citealt{iyyer2015deep}) as the student model in this paper. 
    DAN contains a sparse n-gram embedding table and two linear layers.
    Embedding dimension $d_e$ is set to 3 in this figure for illustration purpose. 
    }
    \label{fig:dan}
\end{figure}

Counter to the convention of using ``smaller, faster, [and] cheaper'' \citep{sanh2019distilbert} student models, our work explores a new area of the design space, where our fast and cheap student model is actually several times \textit{larger} than the teacher. The student model we use is modified from Deep Averaging Network (DAN) in \citet{iyyer2015deep}. DANs take a simple architecture by mapping the n-grams in the input sentence into embeddings, aggregating the embeddings with average pooling, and then using multiple linear layers to perform classification (see Fig.~\ref{fig:dan}). This architecture is reminiscent of the high expressive power of billion-parameter n-gram models \citep{buck2014n,brants2007large} from before the existence of pre-trained language models. By selecting the n-gram vocabulary and the embedding dimension, DANs also scale up to billions of parameters. Meanwhile, the inference costs are kept low as DANs are sparsely-activated. 

One weakness of DANs is that they are restricted in modeling high-level meanings in long-range contexts, as compared to the self-attention operator in Transformers. However, recent studies have shown that large pre-trained Transformers are rather insensitive to word order \citep{sinha2021masked} and that they still work well when the learned self-attention is replaced with hard-coded localized attention \citep{you2020hard} or convolution blocks \citep{tay2021pre}. Taken together, these studies suggest that on some tasks it may be possible to get competitive results without computationally expensive operations such as self-attention.

To verify our hypothesis, we use six single-sentence text classification tasks\footnote{Transformers are effective at many tasks beyond text classification. We extend our method to sentence-pair tasks in later sections and leave other use cases as future work.} and apply knowledge distillation to DANs. 
We observe that the resulting student models retain 97\% of the RoBERTa-Large teacher performance on average.
We also show that our method falls outside of the Pareto frontier of existing methods; compared to a baseline of distilling to a LSTM student, our method gives comparable accuracy at less than 1/20 the inference cost (see Fig.~\ref{fig:speed}). 
Based on our empirical results, we conclude that faster and larger student models provide a valuable benefit over existing methods. We further examine our method (1) with QQP, a sentence-pair task, (2) in privacy-preserving settings (\textit{i.e.}, no access to task-specific data during distillation), and (3) in domain generalization and adaptation settings (\textit{i.e.}, student models are applied and adapted to new data domains), where we find our method continues to bring improvements over non-distillation baselines.
  

\section{Sparse Distillation with DANs}
\subsection{Problem Definition} 
Our goal is to train an \textbf{efficient} text classification model $M$ for a given task $T$. In a $n$-way classification problem, the model $M$ takes input text $\mathbf{x}$, and produces $\hat{\mathbf{y}}\in\mathbb{R}^{n}$, where $\hat{y}_i$ indicates the likelihood that the input $\mathbf{x}$ belongs to category $i$. The task $T$ has a train set $D_{train}$ and a validation/development set $D_{dev}$. Additionally, we assume access to a large unlabeled corpus $C$ which is supposedly in a domain relevant to task $T$. We comprehensively evaluate the efficiency of the model $M$ by reporting: (1) accuracy on $D_{dev}$, (2) inference speed, and (3) the number of parameters in the model.

\begin{figure*}
    \centering
    \includegraphics[width=\textwidth]{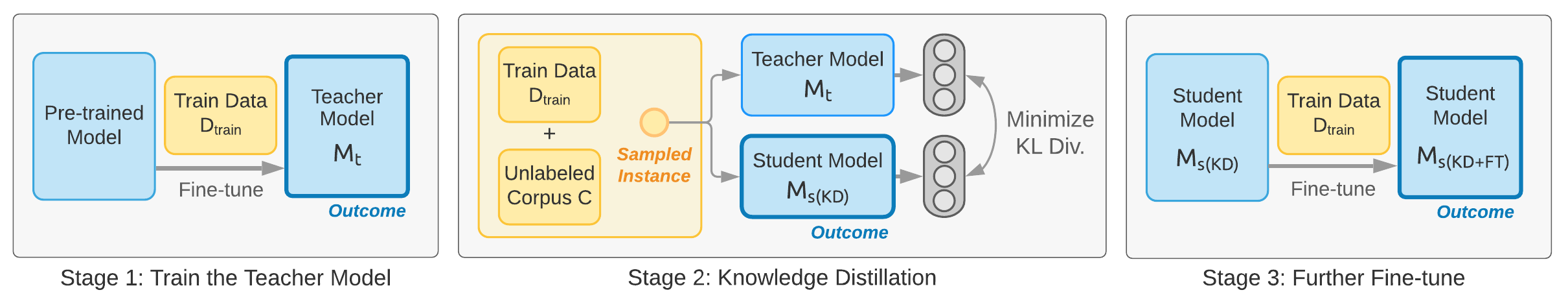}
    \caption{We adopt a \textbf{three-stage pipeline} for Sparse Distillation: (1) We fine-tune a RoBERTa-Large model on $D_{train}$ to get the teacher model. (2) We apply teacher model to the unlabeled corpus $C$ and $D_{train}$, and train the student model (DAN) to mimic the predictions of the teacher. This model is denoted as ``DAN (KD)'' (3) We further fine-tune the student model with $D_{train}$. This model is denoted as ``DAN (KD+FT)''.}
    \label{fig:pipeline}
\end{figure*}

\subsection{Method Overview}
To train a text classifier that is both efficient and powerful, we employ knowledge distillation \cite{hinton2015distilling}, by having a powerful teacher model provide the supervision signal to an efficient student model. In particular, we are interested in using sparse n-gram based models as our student model. We explain the teacher and student model we use in \S\ref{ssec:model}, the training pipeline in \S\ref{ssec:pipeline}, and implementation details in \S\ref{ssec:training_details}

\subsection{Models}\label{ssec:model}
\paragraph{Teacher Model.} Fine-tuning a pre-trained transformer model is the predominant recipe for obtaining state-of-the-art results on various text classification tasks. Our teacher model is a RoBERTa-Large model \citep{liu2019roberta} fine-tuned on the training set $D_{train}$ of task $T$. 

\paragraph{Student Model.} 
Our student model is based on the Deep Averaging Network (DAN, \citealt{iyyer2015deep}) with the modification that we operate on \textit{n-grams} instead of just \textit{words}. See Fig.~\ref{fig:dan} for an illustration of the model architecture.  
Specifically, for an input sentence $\mathbf{x}$, a list of n-grams ${g_1, g_2, ..., g_n}$ are extracted from the sentence. These n-gram indices are converted into their embeddings (with dimension $d_e$) using an embedding layer $\text{Emb}(.)$.
The sentence representation $\mathbf{h}$ will be computed as the average of all n-gram embeddings, \textit{i.e.}, $\mathbf{h}=\text{Mean}(\text{Emb}(g_1), \text{Emb}(g_2), ..., \text{Emb}(g_n))\in\mathbb{R}^{d_e}$. The sentence representation then goes through two fully connected layers, $(\mathbf{W}_1, \mathbf{b}_1)$ and $(\mathbf{W}_2, \mathbf{b}_2)$, to produces the final logits $\hat{\mathbf{z}}$, i.e., $\hat{\mathbf{z}}=M_s(\mathbf{x})=\mathbf{W}_2 (\text{ReLu}(\mathbf{W}_1 \mathbf{h} + \mathbf{b}_1))+\mathbf{b}_2\in\mathbb{R}^{n}$. The logits are transformed into probabilities with the Softmax function, i.e., $\hat{\mathbf{y}}=\text{Softmax}(\hat{\mathbf{z}})\in\mathbb{R}^{n}$. 

\paragraph{Remarks on Computation Complexity.}
Multi-headed self-attention is considered the most expensive operation in the teacher transformers, where the computation complexity is $O(m^2)$ for a sequence with $m$ sub-word tokens.
The student model, Deep Averaging Network (DAN), can be considered as pre-computing and storing phrase representations in a large embedding table. By doing so, the computation complexity is reduced to $O(m)$. However, unlike the teacher, the context is limited to a small range, and no long-range information (beyond n-gram) is taken into account by the student model.


\subsection{Training Pipeline}
\label{ssec:pipeline}
Our training pipeline is illustrated in Fig.~\ref{fig:pipeline}.
It has three stages: 
\textbf{(1)} We first fine-tune a RoBERTa-Large model on the train set $D_{train}$ of task $T$, and use the resulting model as the teacher model. 
\textbf{(2)} We train the student model by aligning the predictions of the teacher ($\tilde{y}$) and the predictions of the student ($\hat{y}$) on the union of unlabeled corpus $C$ and the train set $D_{train}$. We align the predictions by minimizing the KL divergence between the two distributions, \textit{i.e.}, $L=\sum_{j=1}^n \tilde{y}_j \log \frac{\tilde{y}_j}{\hat{y}_j}$. The resulting student model is denoted as ``DAN (KD)''.
\textbf{(3)} We further fine-tune the student model from step (2) with the task train set $D_{train}$, and get a new student model. This model is denoted as ``DAN (KD+FT)''. 
This third stage is optional.


\subsection{Implementation Details}
\label{ssec:training_details}


\paragraph{Determine N-gram Vocabulary.}
Our student model takes in n-grams as input. We determine the n-gram vocabulary by selecting the top $|V|$ frequent n-grams in $D_{train}$ and $C$. For each downstream dataset, we compute the vocabulary separately. We use \texttt{CountVectorizer} with default whitespace tokenization in \texttt{sklearn} \citep{scikit-learn} to perform this task. We set n-gram range to be $(1,4)$ and set \mbox{$|V|= $ 1,000,000}, $d_e=1,000$, unless specified otherwise. 

\paragraph{Optimization.} 
The architecture of DAN is sparsely-activated, and thus can be sparsely-optimized to reduce memory footprint. 
To facilitate this, we design a hybrid Adam optimizer, where we use SparseAdam\footnote{Source code: \url{https://pytorch.org/docs/master/generated/torch.optim.SparseAdam.html}. Please refer to Appendix~\ref{app:adam} for a brief introduction on SparseAdam.} for the sparse parameters (\textit{i.e.}, the embedding layer), and regular Adam for dense parameters.
This implementation helps to improve speed and reduce memory usage greatly~--~we can train a 1-billion parameter DAN with the batch size of 2048 at the speed of 8 batches/second, on one single GPU with 32 GB memory.

\paragraph{Additional Details.} Due to space limit, we defer details such as hyper-parameters settings and hardware configurations in Appendix~\ref{app:reproduce}.




\section{Experiment Settings}

\subsection{Data}

\textbf{Downstream Datasets.}
Following \citet{tay2021pre}, we mainly use six single-sentence classification datasets as the testbed for our experiments and analysis. These datasets cover a wide range of NLP applications.
We use IMDB \citep{maas-etal-2011-learning} and SST-2 \citep{socher-etal-2013-recursive} for sentiment analysis, TREC \citep{li-roth-2002-learning} for question classification, AGNews \citep{zhang2015character} for news classification. 
We use Civil Comments \citep{borkan2019nuanced} and Wiki Toxic \citep{wulczyn2017} dataset for toxicity detection.

\noindent\textbf{Knowledge Distillation Corpora.}
We manually select a relevant unlabeled corpus $C$ based on the task characteristics and text domain.\footnote{It is possible that a careful comparison of different distillation corpora can result in better performance. For the purpose of this study, we leave this as future work.} 
For example, the IMDB and SST-2 models, which are tasked with classifying the sentiment of movie reviews, are paired with a corpus of unlabeled Amazon product reviews \cite{ni2019justifying}. 
TREC, a question classification task, is paired with PAQ \cite{lewis2021paq}, a collection of 65 million questions. 
AGNews, a news classification task, is paired with CC-News corpus \cite{nagel2016cc}. 
For Civil Comments, a dataset for detecting toxic news comments, we select the News subreddit corpus from ConvoKit \cite{chang2020convokit}, which is built from a previously existing dataset extracted and obtained by a third party and hosted by \url{pushshift.io}.
Details of all datasets and corpora are listed in Table \ref{table:datasets}.

\begin{table}[t]
\begin{center}
\scalebox{0.62}{
\begin{tabular}{lrrrlr}
\toprule
Dataset $D$        & $|D_{train}|$ & $|D_{dev}|$ & Avg. $l$ & Distillation Corpus $C$ & $|C|$                \\ \midrule
IMDB           & 25,000   & 25,000  & 300 & Amazon Reviews and $\star$                                      & 75m \\
SST-2          & 67,349   & 872    & 11 & Amazon Reviews    & 75m  \\
TREC           & 5,452    & 500    & 11 & PAQ                                                                     & 65m                  \\ 
AGNews         & 120,000  & 7,600   & 55 & CC-News                                                                 & 418m                 \\ 
CCom        & 1,804,874 & 97,320  & 67 & Reddit News and $\star$ & 60m     \\
WToxic     & 159,571  & 63,978  & 92 & $\star$  & 37m                  \\
\bottomrule

\end{tabular}
}
\caption{\textbf{Datasets and Distillation Corpus Used in Our Study.} $|.|$ represents the size of a dataset. ``Avg. $l$'' represents the average number of tokens in the input sentence. $\star$ represents the unlabeled data released with the original dataset.}\label{table:datasets}
\end{center}
\end{table}

\subsection{Compared Methods}\label{ssec:baselines}
To comprehensively evaluate and analyze the n-gram student models, we additionally experiment with (1) training a randomly-initialized DAN model with $D_{train}$, without knowledge distillation (``from scratch''); (2) directly fine-tuning general-purpose compact transformers, \textit{e.g.}, DistilBERT \citep{sanh2019distilbert}, MobileBERT \citep{sun2020mobilebert}; (3) using other lightweight architectures for the student model, such as DistilRoBERTa \citep{sanh2019distilbert}, Bi-LSTM \cite{tang2019distilling} and Convolution Neural Networks \cite{chia2019transformer}, in task-specific distillation setting. We also quote performance from \cite{tay2021pre} when applicable.








\begin{table*}[t]
\centering
\scalebox{0.7}{
\begin{tabular}{l|C{0.08\textwidth}C{0.08\textwidth}C{0.08\textwidth}C{0.1\textwidth}C{0.08\textwidth}C{0.09\textwidth}|C{0.08\textwidth}}
\toprule
Model & IMDB & SST-2 & TREC & AGNews & CCom & WToxic & QQP\\\midrule
DAN (from scratch) & 88.3 & 79.5 & 78.4 & 91.1 & 95.7 & 92.2 & 82.0\\
DAN (KD)$^\dagger$ & 92.0 & 87.0  & 91.8 & 90.0 & 96.2 & 93.9 & 63.2 \\
DAN (KD) & 93.2 & 86.4 & 91.8 & 90.6 & 96.3 & 94.0 & 84.1\\
DAN (KD+FT) & 93.5 & 88.5 & 92.6 & 93.0 & 96.3 & 92.5 & 84.2\\ 
\midrule
DistilBERT \cite{sanh2019distilbert} & 92.2 & 90.8 & 92.8 & 94.5 & 96.9 & 93.1 & 89.4 \\
MobileBERT \cite{sun2020mobilebert} & 93.6 & 90.9 & 91.0 & 94.6 & 97.0& 93.5 & 90.5 \\\midrule
Transformer-Base \citep{tay2021pre} & 94.2 & 92.1 &  93.6 & 93.5 & -$^\ddag$ & 91.5 & - \\
ConvNet \citep{tay2021pre} & 93.9 & 92.2 & 94.2 & 93.9 & -$^\ddag$ & 93.8 & - \\
RoBERTa-Large \citep{liu2019roberta} & 96.3 & 96.2 & 94.8 & 95.4 & 96.3 & 94.1 & 92.1 \\
\bottomrule
\end{tabular}
}
\caption{\textbf{Performance Comparison on 6 Single-sentence Tasks and 1 Sentence-pair Task.} We report accuracy for all datasets. For single-sentence tasks, the gap between the teacher model (RoBERTa-Large) and the n-gram based student model (DAN(KD)/DAN(KD+FT)) is within 3\% in most cases. Also, we observe that knowledge distillation help close more than half the gap between the teacher model and the n-gram model trained from scratch. $^\dagger$Knowledge distillation is performed without task data ($D_{train}$), assuming that the task data is private (see \S\ref{ssec:domain}). $^\ddag$The dataset we obtain from public sources differs from the one in \citet{tay2021pre}.
}
\label{table:main}
\end{table*}
\begin{table*}[t]
\centering
\scalebox{0.7}{
\begin{tabular}{l|cccccc|c}
\toprule
Model & IMDB & SST-2 & TREC & AGNews & CCom & WToxic & QQP\\\midrule
RoBERTa-Large & 29 (1x) & 298 (1x) & 549 (1x) & 147 (1x) & 35 (1x) & 72 (1x) & 240 (1x) \\
DistilBERT & 176 (6x) & 1055 (4x) & 930 (2x) & 740 (5x) & 188 (5x) & 426 (6x) & 1201 (5x) \\
MobileBERT & 158 (5x) & 736 (3x) & 402 (1x) & 751 (5x) & 187 (5x) & 400 (6x) & 943 (4x) \\
DANs & 17557 (607x) & 3020 (10x) & 2236 (4x) & 24084 (164x) & 38024 (1091x) & 48133 (668x) & 35708 (149x) \\
\bottomrule
\end{tabular}
}
\caption{\textbf{Inference Speed Comparison (Unit: samples per second).} DANs greatly improves inference speed, with the speed-up ranging from 4x to 1091x. Speed-up is most significant with classification tasks with long sequences as input, \textit{e.g.}, Civil Comment, Wiki Toxic, and IMDB.
}
\label{table:speed2}
\end{table*}
\section{Results and Analysis}

\subsection{Main Results}
\paragraph{How well can DANs emulate the performance of the teacher?}
\label{ssec:main_results}
In Table~\ref{table:main}, we present the results on 6 single-sentence classification datasets. 
Firstly, we find that in 5 out of the 6 datasets, the gap between the teacher and the student model is within 3\%. 
This suggests the power of simple n-gram models may be underestimated previously, as they are typically trained from scratch, without modern techniques such as pre-training and knowledge distillation. This also echoes with a series of recent work that questions the necessity of word order information \cite{sinha2021masked} and self-attention \cite{you2020hard}, in prevalent transformer architectures.
Secondly, we observe that knowledge distillation help close more than half the gap between the teacher model and the student model trained from scratch. The effect is more significant with TREC dataset (13\% improvement), a 46-way classification problem, whose train set has a small size of 5,452. It is hard to estimate parameters of a large sparse model with merely 5,452 examples; however, supervising it with large-scale corpus and distillation target effectively densified the supervision signals and help address the sparsity issues during model training. 

\paragraph{How fast are DANs?}
We have previously hypothesized that DANs will have superior inference speed due to its simple and sparse architecture. In this section we quantify this advantage by comparing the student model with the RoBERTa-Large teacher model. We also include the baselines listed in \S\ref{ssec:baselines} for a comprehensive comparison. For simplicity, we use BPE tokenizer and re-use the embedding table from RoBERTa-Large for our student Bi-LSTM and CNN model. We use 2-layer Bi-LSTM with hidden dimension of 4, 64, 256 and 512. For the CNN model, we use one 1D convolution layer with hidden dimension of 128 and context window of 7.

We provide speed comparison across all datasets in Table~\ref{table:speed2}. We provide more fine-grained comparison on IMDB dataset in Table~\ref{table:speed_lstm} and Fig.~\ref{fig:speed}.
DAN achieves competitive performance and the fastest inference efficiency among all different student model architectures.
The speed-up differs across datasets, ranges from 4x to 1091x. It is most significant on Civil Comments (1091x), Wiki Toxic (668x) and IMDB dataset (607x), as they have longer input sequences, and the complexity grows quadratically with sequence length in transformer models.
Moreover, as shown in Table~\ref{table:speed_lstm}, DAN has an acceptable CPU inference speed, which greatly reduce the hardware cost for inference.
We believe all these characteristics makes student DAN model as an ideal option for production or real-time use on single-sentence classification tasks.

\begin{table}[t]
\centering
\scalebox{0.6}{
\begin{tabular}{l|c|crr}
\toprule
& \multicolumn{1}{c|}{Parameter Count} & \multicolumn{3}{c}{IMDB} \\
                   & Total/Sparse/Dense & Acc. & GPU Speed & CPU Speed   \\\midrule
\ \ RoBERTa-Large      & 355M/51M/304M           & 96.3     & 29 (1x)  & 1 (1x)  \\
\ \ DistilBERT  & 66M/23M/43M & 92.2 & 176 (6x)  & 11 (8x) \\
\ \ MobileBERT  & 25M/4M/21M & 93.6 & 158 (5x)  & 8 (6x) \\
$\star$DistilRoBERTa & 83M/39M/44M & 95.9 & 176 (6x)  & 8 (6x) \\
\midrule
$\star$LSTM (2l-512d) & 62M/51M/11M & 95.9     & 362 (12x)  & 31 (22x) \\
$\star$LSTM (2l-256d) & 56M/51M/5M  & 95.8     & 665 (23x)  & 52 (37x) \\
$\star$LSTM (2l-64d) & 53M/51M/2M   & 95.3     & 818 (28x)  & 101 (73x) \\
$\star$LSTM (2l-4d) & 52M/51M/<1M    & 93.1     & 813 (28x)  & 146 (105x) \\
$\star$CNN (1l-256d)  & 53M/51M/2M  & 89.2     & 3411 (109x)  & 251 (181x) \\ \midrule
$\star$DAN (this work)     & 1001M/1000M/1M  & 93.5     & 17558 (607x) & 923 (663x)  \\\bottomrule
\end{tabular}
}
\caption{\textbf{Detailed Inference Speed Comparison on IMDB.} DANs achieves better accuracy and inference speed compared to other lightweight architectures such as LSTMs and CNNs. Moreover, DANs achieves acceptable inference speed on CPUs. $\star$ indicates the model is trained with task-specific distillation; no $\star$ indicates the model is trained with direct fine-tuning.
}
\label{table:speed_lstm}
\end{table}


\begin{table}[t]

\centering
\scalebox{0.62}{
\begin{tabular}{lr|lr}
\toprule
Variations & Acc. & Variations & Acc. \\\midrule
\rowcolor{gray!20}\multicolumn{2}{l|}{\textit{1. Pooling Methods}} & \multicolumn{2}{l}{\textit{2. Dense Layers}}\\\midrule
Mean Pooling ($\star$) & 93.2 & $1000\rightarrow1000\rightarrow2$ ($\star$) & 93.2\\
Max Pooling & 91.8 & $1000\rightarrow1000\rightarrow256\rightarrow2$ & 93.1\\
Attentive Pooling & 93.0 & $1000\rightarrow1000\rightarrow256\rightarrow64\rightarrow2$ & 93.0 \\
Sum & 92.9 &\\\midrule
\rowcolor{gray!20}\multicolumn{2}{l|}{\textit{3. Embedding Initialization}} & \multicolumn{2}{l}{\textit{4. Parallel Training}}\\\midrule
Without initialization ($\star$) & 93.2 & 1 GPU, param. 1b ($\star$) & 93.2\\
With initialization & 93.2 & 2 GPUs, param. 2b & 93.1\\
\bottomrule
\end{tabular}
}
\caption{\textbf{Variations made to the student model and the performance on IMDB.} $\star$ represents the design we adopt in our main experiments.}\label{tab:ablation}
\vspace{-0.2cm}
\end{table}

\paragraph{Simplest is the best: Exploring different design choices for DAN.} We try several modifications to our current experiment pipeline, including (1) replace average pooling with max pooling, attentive pooling, or taking sum in the DAN model; (2) pre-compute a n-gram representation by feeding the raw n-gram text to a RoBERTa-Large model, and using the representations to initialize the embedding table of the student model; (3) attach more dense layers in the DAN; (4) use even larger student models by leveraging parallel training across multiple GPUs. More details about these variations are in Appendix~\ref{app:ablation}. We experiment with IMDB dataset and list the performance in Table~\ref{tab:ablation}. In general, we do not observe significant performance improvements brought by these variations. Thus, we keep the simplest design of DAN for all other experiments.

\subsection{Controlling the Parameter Budget}\label{ssec:budget}
Given a fixed parameter budget, how to allocate it wisely to achieve optimal performance? We discuss this question in two scenarios: the users wish to control the parameter budget (1) during knowledge distillation (KD), or (2) during inference.

\paragraph{During KD: Trade-off between vocabulary size and embedding dimension.}
We explore how the configuration of vocabulary size and embedding dimension influence the student model performance.
We train student models on the IMDB dataset with 19 configurations, and show the results graphically in Figure \ref{fig:vocab_tradeoffs}. Detailed results are deferred in Table~\ref{tab:tradeoff} in Appendix~\ref{app:additional}.
All else being equal, having more parameters in the student model is beneficial to the performance. For a fixed parameter budget, higher accuracy was achieved by increasing the embedding dimension and making a corresponding reduction in the vocabulary size. Our best performing model has $|V|=1,000,000$ and $d_e=1,000$. We keep this configuration for the main experiments in previous sections.

\begin{figure}[t]
    \centering
    \vspace{-0.2cm}
    \includegraphics[width=0.4\textwidth]{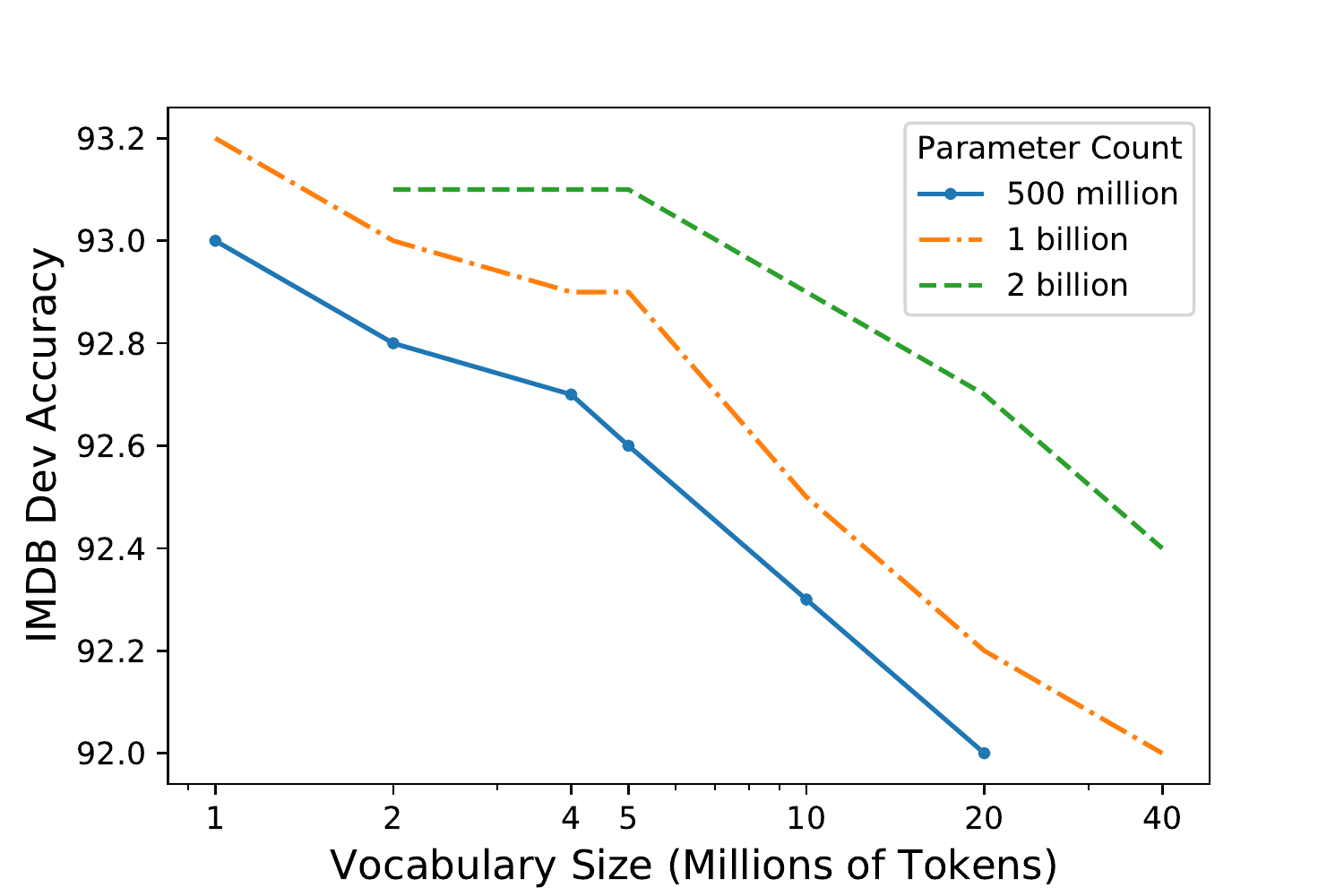}
    \vspace{-0.2cm}
    \caption{\textbf{Trade-off between the vocabulary size and the embedding dimension.} Given a fixed parameter budget, empirical results suggest that a larger embedding dimension and a smaller vocabulary size should be selected.}
    \label{fig:vocab_tradeoffs}
\end{figure}

\begin{figure}[t]
    \centering
    \includegraphics[width=0.48\textwidth]{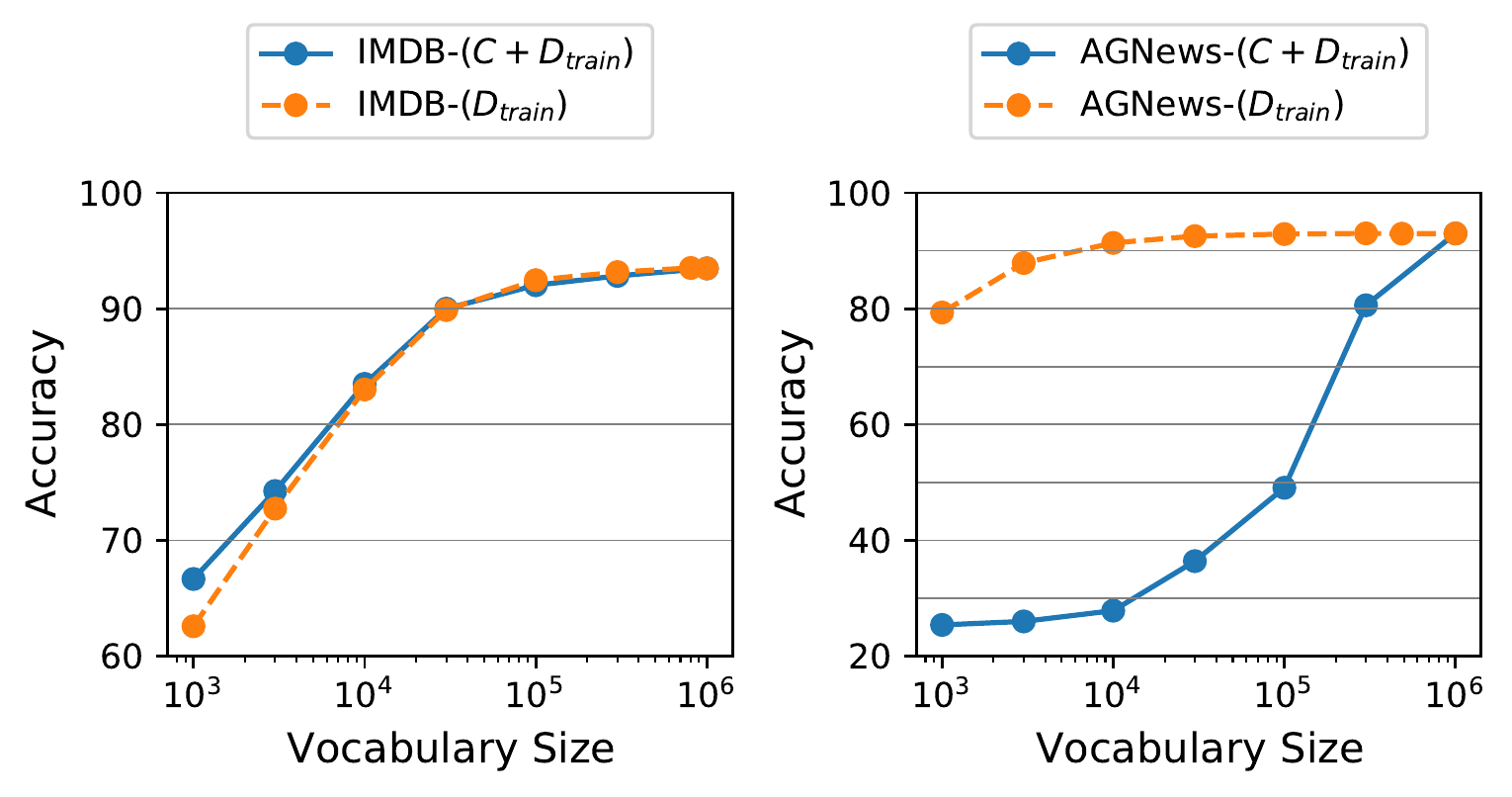}
    \caption{\textbf{Post-hoc pruning according to n-gram frequency.} 
    We disable the \textit{least} frequent n-grams during inference to further reduce model size. 
    When the n-gram frequencies are estimated appropriately, DANs maintain decent performance (acc.>90\%) even when model is 3\% of its original size.
    $C+D_{train}$/$D_{train}$ represent different ways to estimate n-gram frequencies.
    }
    \label{fig:pruning}
\end{figure}

\paragraph{During inference: Reduce the model size with n-gram pruning.}
The model size of DANs is flexible even after training, by excluding the \textit{least} frequent n-grams in the vocabulary. We test this idea on IMDB and AGNews dataset and plot the performance in Fig.~\ref{fig:pruning}. We try two ways to estimate n-gram frequency: (1) using distillation corpus $C$ and the training set $D_{train}$; (2) using $D_{train}$ only. We observe that: (1) n-gram frequencies estimated on $D_{train}$ are more reliable, as $D_{dev}$ has a n-gram distribution more similar to $D_{train}$ compared to $C+D_{train}$; (2) DANs maintain decent accuracy (>90\%) even when the model size is cut to 3\% of its original size. In this case, users of DANs can customize the model flexibly based on their needs and available computational resources.


\subsection{Privacy-preserving Settings}
\label{ssec:domain}

NLP datasets sometimes involve user generated text or sensitive information; 
therefore, data privacy can be a concern when training and deploying models with certain NLP datasets.
In this section, we modify our experiment setting to a practical and privacy-preserving one. 
We assume the user has access to a public teacher model that is trained on private train dataset ($D_{train}$), but does not has access to $D_{train}$ itself. This is realistic nowadays with the growth of public model hubs such as TensorFlow Hub\footnote{\url{https://www.tensorflow.org/hub}} and Hugging Face Models\footnote{\url{https://huggingface.co/models}}.
After downloading the model, the user may wish to deploy a faster version of this model, or adapt the model to the user's own application domain.

\paragraph{Knowledge Distillation without $D_{train}$.} 
To simulate the privacy-preserving setting, we remove $D_{train}$ from the knowledge distillation stage in our experiment pipeline and only use the unlabeled corpus $C$. 
We use ``DAN (KD)$^\dagger$'' to denote this model in Table~\ref{table:main}. By comparing ``DAN (KD)'' and ``DAN (KD)$^\dagger$'', we found that the performance difference brought by task specific data $D_{train}$ is small for all single-sentence tasks, with the largest gap being $1.2\%$ on IMDB dataset.
This suggests that the proposed pipeline is still useful in the absence of task-specific data.

\paragraph{Domain Generalization and Adaptation.} We select the two sentiment analysis tasks: IMDB and SST-2, and further explore the domain generalization/adaptation setting. Specifically, during stage 1 of our three-stage pipeline (\S\ref{ssec:pipeline}), we fine-tune the RoBERTa-Large model on a source dataset; during stage 2, we apply knowledge distillation with unlabeled corpus $C$ only and get the student model; during stage 3, we further fine-tune the student model on the target dataset. The last step is optional and serves to simulate the situation where the user collects additional data for domain adaptation.
We list the results in Table~\ref{tab:generalization}. With weakened assumptions about the teacher model and distillation supervision, we still have observations similar to those in our main experiments (\S\ref{ssec:main_results}): Performance of the final student model is significantly improved compared to DANs trained from scratch. 
\begin{table}[t]
\centering
\scalebox{0.75}{
\begin{tabular}{lcccccc}
\toprule
Source   & IMDB & SST-2 \\
Target   & SST-2   & IMDB     \\\midrule
DAN (from scratch, tar)   & 79.5 & 88.3 \\ \midrule
(1) RoBERTa-Large (src) & 90.0 & 94.1   \\
(2) DAN (KD)        & 81.9 & 92.0  \\
(3) DAN (KD+FT)   &  88.4 & 93.0 \\
(4) DAN (KD+FT w. re-init.) & 86.7 & 92.8 \\\midrule
RoBERTa-Large (tar)      & 96.2 & 96.3 \\\bottomrule
\end{tabular}
}
\caption{\textbf{Domain Generalization and Adaptation Results.} (1) We take the teacher model trained on the \textit{source dataset} and evluate it on the \textit{target dataset}. (2) We obtain the  student model ``DAN (KD)'' with {unlabeled corpus} $C$ and knowledge distillation. (3) We further fine-tune the student model on the \textit{target dataset} to obtain ``DAN (KD+FT)''. (4) The classification head is re-initialized before further fine-tuning.}
\label{tab:generalization}
\end{table}


\subsection{Limitations and Discussions}\label{ssec:limitation}

\begin{table*}[t]
\centering
\scalebox{0.7}{
\begin{tabular}{llll}
\toprule
Teacher & Student & Label & Sentence                     \\\midrule
Negative   & Positive   & Negative          & I really wanted to love this film. \dots                                                             \\
Negative   & Positive   & Negative          & This movie is a great movie ONLY if you need something to sit and laugh at the stupidity of it. \dots \\
Positive   & Negative   & Positive          & \dots They are such bad actors and it made this movie so much funnier to watch. \dots                \\\bottomrule
\end{tabular}
}
\caption{\textbf{Case study on IMDB predictions.} In these cases, the model can only make the correct predictions by understanding long contexts. Performance of DAN models are still limited as they only look at local n-grams.
}\label{tab:case_study}
\end{table*}
\paragraph{Extension to sentence-pair tasks.}
So far we have limited the scope to single-sentence classification tasks. We consider extending our sparse distillation framework to a sentence-pair task, Quora Question Pair (QQP)\footnote{\url{https://quoradata.quora.com/First-Quora-Dataset-Release-Question-Pairs}}, which aims to identify duplicated questions. We create pseudo sentence-pair data for knowledge distillation by randomly sampling 10 million question pairs from PAQ.
To better model the relation between a pair of sentence, we modify DANs by introducing a concatenate-compare operator \cite{wang2016compare}, following the practice in \cite{tang2019distilling}. More specifically, the two input sentences, $\mathbf{x}_1$ and $\mathbf{x}_2$, go through the embedding layer and average pooling independently, resulting in two sentence representations, $\mathbf{h}_1$ and $\mathbf{h}_2$. We then apply the concatenate-compare operator, i.e., $f(\mathbf{h}_1, \mathbf{h}_2)=[\mathbf{h}_1, \mathbf{h}_2, \mathbf{h}_1 \odot \mathbf{h}_2, |\mathbf{h}_1-\mathbf{h}_2|]$, where $\odot$ represents element-wise multiplication. Finally, $f(\mathbf{h}_1, \mathbf{h}_2)$ go through two fully connected layers for classification, the same as DANs for single-sentence tasks. 

The results on QQP dataset is listed in the rightmost column in Table~\ref{table:main}. Firstly, knowledge distillation still helps close the gap between RoBERTa-Large and DANs trained from scratch (2\% improvement) and leads to a decent accruacy of 84.2\%; however the benefit brought by KD is not as strong as with single-sentence tasks. Secondly, the performance of DAN(KD)$^\dagger$ (\textit{i.e.}, without access to $D_{train}$ during KD) is much worse than the performance of DAN(KD). We hypothesize that this is due to the quality and distribution of knowledge distillation corpus. We randomly sample questions pairs as the knowledge distillation examples, which may not carry sufficient supervision signals~--~more than 99\% of them are negative (``not duplicated'') examples. Creating more suitable distillation corpus for sentence-pair tasks is beyond the scope of our work, and we leave this as future work.

\paragraph{Impact of N-gram Coverage.} 

\begin{figure}[t]
    \centering
    \includegraphics[width=0.48\textwidth]{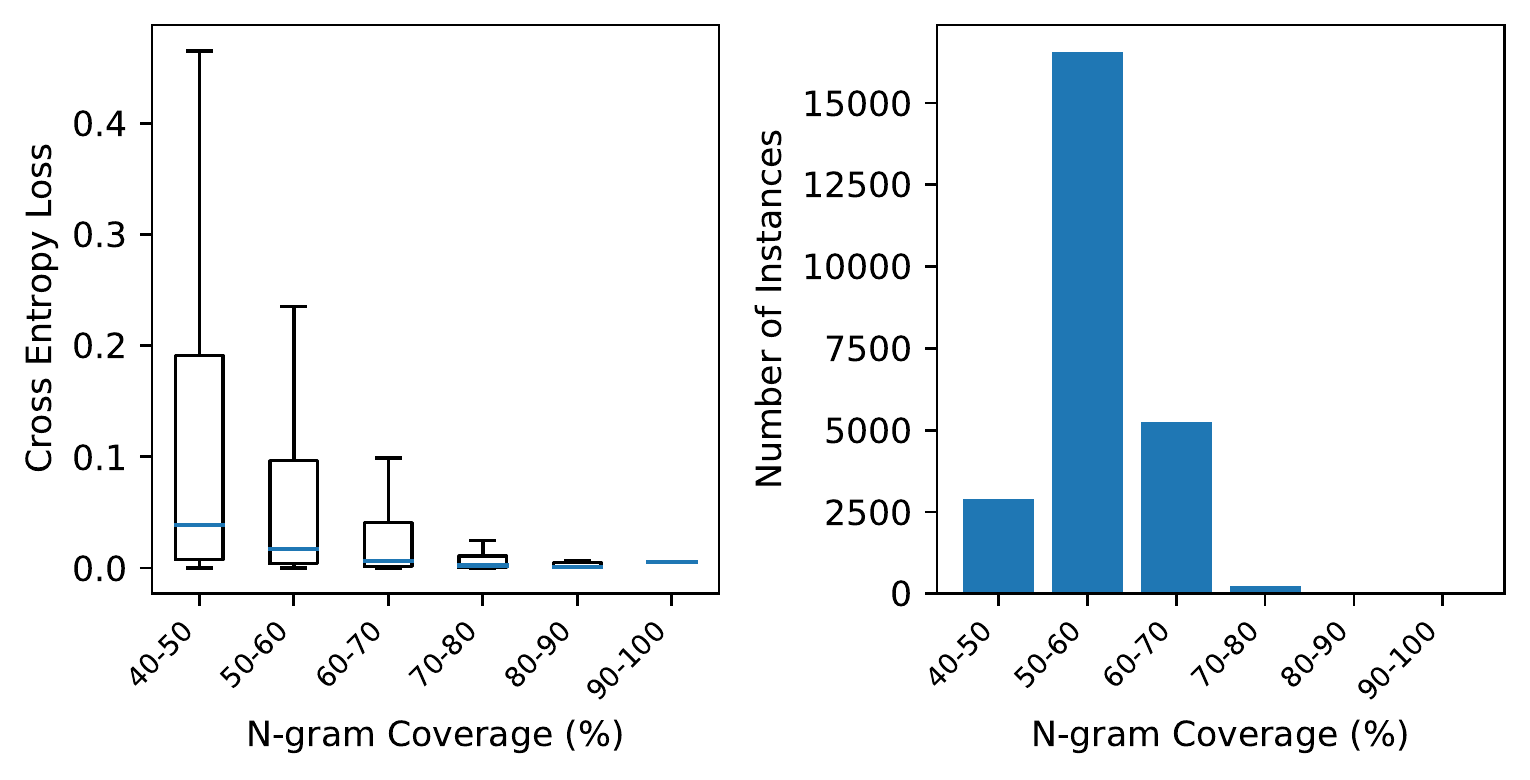}
    \caption{\textbf{Analysis on N-gram Coverage.}
    Left: Relation between n-gram coverage and cross-entropy loss w.r.t. ground truth labels. Each blue line represents the median loss in that n-gram coverage bucket.
    Right: Distribution of n-gram coverage.
    }
    \label{fig:coverage}
\end{figure}
One potential drawback of n-grams (based on white-space tokenization) is that they cannot directly handle out-of-vocabulary words, while WordPiece/BPE tokenization together with contextualization can better handle this issue. In Fig.~\ref{fig:coverage}, we quantify the influence of n-gram coverage on IMDB dev set. 
Here, n-gram coverage for an input sentence is defined as $|G \cap V|/|V|$, where $G$ represents the set of n-grams in the sentence and $V$ is the n-gram vocabulary (\S\ref{ssec:training_details}).
We first group the instances into buckets of n-gram coverage (\textit{e.g.}, $[40\%, 50\%), [50\%,60\%)$) and then compute the statistics of cross-entropy loss in each bucket. We observe that performance is worse on sentences with more out-of-vocabulary words. Future work may build upon this observation and improve DANs performance by addressing out-of-vocabulary words. For example, BPE-based n-grams may be used for creating the vocabulary.

\paragraph{Case study: What are DANs still not capable of?} 
We take a closer look at the predictions made by our DAN model (student) and the RoBERTa-Large model (teacher) on the IMDB dataset. We list several representative cases in Table~\ref{tab:case_study}. These cases typically require understating of complex language phenomena, such as irony, conditional clauses, and slang. In addition, these phenomena typically occur in contexts longer than 4 words, which DANs are not capable of modeling by design. For example, ``bad actors'' can mean ``good actors'' based on the later context ``much funnier to watch''. We conclude that sparse distillation is not suitable to cases where modeling complex language phenomena has a higher priority than improving inference speed.



\paragraph{Understanding the performance gaps.} \citet{tay2021pre} advocate that architectural advances should not be conflated with pre-training. Our experiments further support this claim, if we consider knowledge distillation as a ``substitute'' for pre-training that provides the student model with stronger inductive biases, and interpret the remaining teacher-student performance gap as the difference brought by architectural advances. On the other hand, we believe the power of DANs are previously undermined due to the challenges in optimizing large sparse models with limited supervision. Our experiments show that knowledge distillation effectively densify the supervision and greatly improve the performance of DANs.

\paragraph{Additional Analysis and Specifications.} 
Due to space limit, we leave some additional analysis and specifications in Appendix \ref{app:additional}.
We discuss tokenization speed (Table~\ref{tab:tokspeed}) and impact of $n$ in n-grams (Table~\ref{tab:n_effect}). We provide more detailed speed comparison in Table~\ref{table:speed_appendix}, model storage and memory usage information in Table~\ref{tab:disk}. We provide fine-grained n-gram coverage information in Table~\ref{tab:ngram-stats}.



\section{Related Work}

\paragraph{Efficient Transformers.} Recent work attempts to improve computation or memory efficiency of transformer models mainly from the following perspectives: (1) Proposing efficient architectures or self-attention variants, \textit{e.g.}, Linformer \citep{wang2020linformer}, Longformer \citep{beltagy2020longformer}. \citet{tay2020efficient} provide a detailed survey along this line of work. (2) Model compression using knowledge distillation, \textit{e.g.}, DistillBERT \citep{sanh2019distilbert}, MobileBERT \citep{sun2020mobilebert}, MiniLM \citep{wang2020minilm}. These compressed models are typically task-agnostic and general-purpose, while in this work we focus on task-specific knowledge distillation. (3) Weight quantization and pruning, \textit{e.g.}, \citet{gordon2020compressing, li2020train, kundu2021attention}.

\paragraph{Task-specific Knowledge Distillation in NLP.} Researchers explored distilling a fine-tuned transformer into the following lightweight architectures, including smaller transformers \citep{turc2019well, jiao2019tinybert}, LSTMs \citep{tang2019distilling, adhikari-etal-2020-exploring} and CNNs \citep{chia2019transformer}.
\citet{wasserblat2020exploring} distill BERT into an architecture similar to DAN, however they restrict the model to only take unigrams (thus having small student models), and adopt a non-standard low-resource setting. To summarize, existing work typically focuses on reducing \textit{both} number of parameter and the amount of computation, while in the paper we study an under-explored area in the design space, where the amount of computation is reduced by training a \textit{larger} student model.

\paragraph{Reducing Contextualized Representations to Static Embeddings.} Related to our work, \citet{ethayarajh2019contextual} and \citet{bommasani2020interpreting} show how static word embeddings can be computed from BERT-style transformer models. \citet{ethayarajh2019contextual} suggest that less than 5\% of the variance in a word's contextualized representation can be explained by a static embedding, justifying the necessity of contextualized representation. \citet{bommasani2020interpreting} found that static embeddings obtained from BERT outperforms Word2Vec and GloVe in intrinsic evaluation. These two papers mainly focus on post-hoc interpretation of pre-trained transformer models using static embeddings.
In our work we opt to use knowledge distillation to learn n-gram embeddings. Meanwhile we acknowledge that 
the technique in \citet{ethayarajh2019contextual} and \citet{bommasani2020interpreting} could be used as an alternative method to convert transformer models to fast text classifiers.


\paragraph{Sparse Architectures.}
In our work we aggressively cut off computation cost by compensating it with more parameters in the student model.
Alternatively, one could fix the computational cost at the same level as a transformer while greatly expanding the parameter count, as explored in the Switch Transformer \citep{fedus2021switch}. Both their work and ours agree in the conclusion that scaling up parameter count allows the model to memorize additional useful information.


\section{Conclusions \& Future Work}

We investigated a new way of using knowledge distillation to produce a faster student model by reversing the standard practice of having the student be smaller than the teacher and instead allowed the student to have a large table of sparsely-activated embeddings. This enabled the student model to essentially memorize task-related information that if an alternate architecture were used would have had to be computed. We tested this method on six single-sentence classification tasks with models that were up to 1 billion parameters in size, approximately 3x as big as the RoBERTa-Large teacher model, and found that the student model was blazing fast and performed favorably.

We hope that our work can lead to further exploration of sparse architectures in knowledge distillation. There are multiple directions for future work, including extending the DAN architecture to better support tasks with long range dependencies like natural language inference or multiple inputs like text similarity. Additionally, more work is needed to test the idea on non-English languages where n-gram statistics can be different from English.

\section*{Acknowledgments} 
We would like to thank Robin Jia, Christina Sauper, and USC INK Lab members for the insightful discussions. We also thank anonymous reviewers for their valuable feedback. Qinyuan Ye and Xiang Ren are supported in part by the Office of the Director of National Intelligence (ODNI), Intelligence Advanced Research Projects Activity (IARPA), via Contract No. 2019-19051600007, the DARPA MCS program under Contract No. N660011924033, the Defense Advanced Research Projects Agency with award W911NF-19-20271, NSF IIS 2048211, NSF SMA 1829268.


\bibliography{iclr2022_conference}
\bibliographystyle{acl_natbib}

\appendix
\section{Reproducibility}
\label{app:reproduce}
\subsection{Datasets}
Datasets and corpora used, and their specifications are previously listed in Table \ref{table:datasets}. Here we provide links to download these data.
\begin{itemize}
    \item IMDB: \url{https://ai.stanford.edu/~amaas/data/sentiment/}
    \item SST-2: \url{https://huggingface.co/datasets/glue}
    \item AGNews: \url{https://huggingface.co/datasets/ag_news}
    \item TREC: \url{https://huggingface.co/datasets/trec}
    \item CivilComments: \url{https://huggingface.co/datasets/civil_comments}
    \item WikiToxic: \url{https://www.tensorflow.org/datasets/catalog/wikipedia_toxicity_subtypes} and \url{https://meta.m.wikimedia.org/wiki/Research:Detox/Data_Release}
    \item QQP: \url{https://huggingface.co/datasets/glue}
    \item Amazon Reviews: \url{https://nijianmo.github.io/amazon/index.html}
    \item PAQ: \url{https://github.com/facebookresearch/PAQ}
    \item Reddit News: \url{https://zissou.infosci.cornell.edu/convokit/datasets/subreddit-corpus/corpus-zipped/newreddits_nsfw~-~news/news.corpus.zip}
\end{itemize}
QQP dataset has 363,846 training instances and 40,430 development instances. The average input length is 13 tokens.
We thank huggingface dataset team \cite{lhoest-etal-2021-datasets} for providing easy access to these datasets. 

\paragraph{Licensing.} For WikiToxic, the dataset is licensed under CC0, with the underlying comment text being governed by Wikipedia's CC-SA-3.0. The PAQ QA-pairs and metadata is licensed under CC-BY-SA. The licensing information of other datasets are unknown to us.

\subsection{Implementation Details}
N-gram pre-processing are implemented with \texttt{scikit-learn} \citep{scikit-learn}. DistilBERT \cite{sanh2019distilbert} and MobileBERT baselines are implemented in \texttt{huggingface transformers} \cite{wolf-etal-2020-transformers}. RoBERTa-Large, BiLSTM, CNN, and DAN experiments are implemented with \texttt{fairseq} \cite{ott-etal-2019-fairseq}. 

\subsection{Hyperparameters} 
For fine-tuning in stage 1, we select batch size from \{16, 32\} and learning rate from $\{$1e-5, 2e-5, 5e-5$\}$ following the recommendations in \citep{liu2019roberta}. We train the model for 10 epochs on $D_{train}$. For knowledge distillation in stage 2, we set the batch size to be 2048, learning rate to be 5e-4, and total number of updates to be 1,000,000, as they work well in our preliminary experiments. The embedding table is randomly initialized and the embedding dimension $d_e$ is set to 1,000, unless specified otherwise. For further fine-tuning in stage 3, we set the batch size to be 32 and select the learning rate from $\{$3e-4, 1e-4, 3e-5$\}$. We train the model for 10 epochs on $D_{train}$. For all training procedures, we validate the model at the end of each epoch in the case of fine-tuning, or every 100,000 steps in the case of knowledge distillation. We save the best checkpoint based on dev accuracy. Due to the analysis nature of this work and the scale of experiments, performance are computed using dev set and based on one single run.

\subsection{Hardware}
\paragraph{Model Training.} Except for the parallel training attempt in Table~\ref{tab:ablation}, all experiments are done on one single GPU. We train DAN models on either A100 40GB PCIe or Quadro RTX 8000 depending on availability. Knowledge distillation (Stage 2) with 1,000,000 updates typically finishes within 36 hours.
\paragraph{Inference Speed Tests.} All inference speed tests are done with the batch size of 32. GPU inference is performed with one Quadro RTX 8000 GPU, and CPU inference is performed with 56 Intel Xeon CPU E5-2690 v4 CPUs.

\section{Additional Details}
\subsection{DAN Variations}
Due to space limits we have omitted the details for the DAN variations we studied in \S\ref{ssec:main_results}. We introduce these variations in the following.
\label{app:ablation}
\paragraph{Attentive Pooling.} We consider adding attentive pooling to the DAN model to capture more complicated relations in the input. Our attention layer is modified from the one in \cite{zhang-etal-2017-position}. we use the representation $\mathbf{h}$ after mean pooling as query, and each n-gram embedding $\mathbf{e}_i=\text{Emb}(g_i)$ as key. More specifically, for each n-gram $g_i$ we calculate an attention weight $a_i$ as:
\begin{align}
    u_i&=\mathbf{v}^\top\tanh(\mathbf{W}_g \mathbf{e}_i+\mathbf{W}_h \mathbf{h})\\
    a_i&=\frac{\exp (u_i)}{\sum_{j=1}^n \exp (u_j)}
\end{align}

Here $\mathbf{W}_g, \mathbf{W}_h\in \mathbb{R}^{d_e\times d_a}$ and $\mathbf{v}\in \mathbb{R}^{d_a}$ are learnable parameters. $d_e$ is the dimension of the embedding table, and $d_a$ is the size of the attention layer. To maintain an acceptable training speed, for attentive pooling, we use a batch size of 512 during knowledge distillation.

\paragraph{Parallel Training}
We try further scaling up the student model by splitting the gigantic embedding table to different GPUs and enable parallel training, as implemented in Megatron-LM \cite{shoeybi2019megatron}. We train a 2-billion parameter model in parallel on two GPUs. The embedding dimension is set to be $2,000$ in total, and each GPU handles an embedding table of hidden dimension $1,000$. The vocabulary size is 1 million.

\subsection{Comments on SparseAdam} 
\label{app:adam}
SparseAdam is a modified version of the regular Adam optimizer. 
For Adam, the first and second moment for \textit{each} parameter is updated at \textit{every} step. 
This can be costly, especially for DAN, as most parameters in the embedding layer are not used during the forward pass. 
SparseAdam computes gradients and updates the moments \textit{only} for parameters used in the forward pass.

\section{Additional Results}
\label{app:additional}
\paragraph{Speed Comparison.}
Table~\ref{table:speed_appendix} is an extended version of Table~\ref{table:speed_lstm} which contains inference speed comparison on IMDB and SST-2 dataset, in three different settings (GPU-FP32, GPU-FP16, CPU-FP32). Our major conclusion remains the same: DANs achieve excellent inference speed in various settings.

\paragraph{Vocabulary Size vs. Embedding Dimension Trade-off.}
Table~\ref{tab:tradeoff} contains original results that were visualized in Fig.~\ref{fig:vocab_tradeoffs}.

\begin{table}[h]
\vspace{0.2cm}
\centering
\scalebox{0.8}{
\begin{tabular}{r|rr|rr|rr}
\toprule
& \multicolumn{2}{c|}{\textit{Param. 500m}} & \multicolumn{2}{c|}{\textit{Param. 1b}} & \multicolumn{2}{c}{\textit{Param. 2b}}\\\midrule
$|V|$  &  $d_e$  & Acc  &  $d_e$  & Acc. & $d_e$  & Acc.      \\\midrule
1m & 500 & 93.0 & \textbf{1000} & \textbf{93.2} & -- & -- \\
2m & 250 & 92.8 & 500 & 93.0 & 900 & 93.1 \\
4m & 125 & 92.7 & 250 & 92.9 & 500 & 93.1 \\
5m & 100 & 92.6 & 200 & 92.9 & 400 & 93.1 \\
10m & 50 & 92.3 & 100 & 92.5 & 200 & 92.9\\
20m & 25 & 92.0 & 50 & 92.2 & 100 & 92.7\\
40m & -- & -- & 25 & 92 & 50 & 92.4\\\midrule
\end{tabular}
}
\caption{IMDB dev accuracy with different configurations of vocabulary size ($|V|$) and embedding table dimension ($d_e$). Performance grows with larger embedding tables, and the best performing model has $|V|=1m$ and $d_e=1,000$.
}
\label{tab:tradeoff}
\end{table}

\paragraph{N-gram Coverage Statistics.} 
In our work, we opt to determine the n-gram vocabulary with the training set $D_{train}$ and the corpus $C$, by selecting the top 1 million n-grams according to frequency. N-gram range is set to be within 1 to 4.
For reference, we list statistics about the n-gram vocabulary in Table~\ref{tab:ngram-stats}. 
It is possible that adjustments to this pre-processing step (\textit{e.g.}, up-weighting n-grams in $D_{train}$ and down-weighting n-grams in $C$) will further improve performance, however we stop further investigation.

\paragraph{Tokenization Speed.} The speed comparison in our work does not take pre-processing process into account. When the inference speed is at millisecond level (\textit{e.g.}, with our DAN model), pre-processing time can become non-negligible. For reference, in Table~\ref{tab:tokspeed} we report the tokenization time on the 25,000 training instances in the IMDB dataset with (1) n-gram tokenization (used by DAN, implemented with \texttt{scikit-learn}); (2) BPE tokenization (used by RoBERTa/DistilRoBERTa, implemented with \texttt{fairseq}); (3) WordPiece tokenization (used by DistilBERT, implemented with \texttt{huggingface transformers}). 

\begin{table}[h]
\centering
\scalebox{0.6}{
\begin{tabular}{lll}
\toprule
Tokenization Method & Time & Complexity \\\midrule
BPE       & 26.46 & $O(n\lg n)$ or $O(|V|n)$ \cite{song-etal-2021-fast}\\
WordPiece & 20.60 & $O(n^2)$ or $O(mn)$ \cite{song-etal-2021-fast}\\
N-gram    & 16.45 & $O(n)$ \\\bottomrule
\end{tabular}
}
\caption{Comparison of tokenization speed and complexity. Time is computed for tokenization the train set of IMDB dataset (25,000 instances) with one single worker. Time is averaged across 5 runs. $n$ represents input length.}\label{tab:tokspeed}
\end{table}

First of all, by setting the number of workers to be equal to the batch size (32) we use in the speed test, the tokenization speed will be 48632 instances/sec (=25000/16.45*32), which is roughly 3x faster than the inference speed. Tokenization speed is non-negligible in this case. Still, the main conclusion from the speed comparison remains the same: DANs are typically 10x-100x faster than the compared models.

Secondly, DAN models still have better tokenization speed than transformer models that use BPE/WordPiece tokenization. 
This is because our DAN model computes n-grams based on whitespace tokenization, which can be done in linear time when the n-gram to id mapping is implemented with a hashmap, \textit{i.e.}, $O(n)$ where $n$ is the input length. BPE/WordPiece tokenization has higher complexity according to \citet{song-etal-2021-fast}. 

We would also like to emphasize that this part is also highly dependent on the design choice and implementation. For example, the user could implement a DAN model with BPE tokenzation. The choice and optimization of tokenization is beyond the scope of this work. 

\paragraph{Impact of $n$ in n-grams.} Similar to the post-hoc pruning experiments in \S\ref{ssec:budget}, we gradually disable the usage of four-grams, trigrams and bigrams at inference time, and report the performance in Table~\ref{tab:n_effect}.

\begin{table}[h]
\centering
\scalebox{0.75}{
\begin{tabular}{l|cc|cc}
\toprule
                & \multicolumn{2}{c|}{IMDB} & \multicolumn{2}{c}{AGNews} \\
                & $|V|$        & Acc.      & $|V|$         & Acc.       \\\midrule
$n=1$           & 54,089        & 74.86     & 81,796         & 91.32      \\
$n\leq 2$       & 446,793       & 92.09     & 541,431        & 92.93      \\
$n\leq 3$       & 835,403       & 93.33     & 882,489        & 93.03      \\
$n\leq 4$ (all) & 1,000,000      & 93.47     & 1,000,000       & 92.99     \\\bottomrule
\end{tabular}
}
\caption{\textbf{Impact of $n$ in n-grams.} We disable usage of longer n-grams in the DAN(KD+FT) model. $|V|$ is the size of the vocabulary after disabling.}\label{tab:n_effect}
\end{table}

\paragraph{Model Storage.} In Table~\ref{tab:disk} we provide more details about the disk space and memory required for using DAN models and the baseline models. Note that the GPU memory listed below is the memory used to load the \textit{static} model. During training, more memory will be dynamically allocated during forward and backward passes. DAN uses smaller memory during training because only a small portion of the parameters are activated and trained (see the last row in Table~\ref{tab:ngram-stats}). In this way we are able to use batch sizes as large as 2048 to train DANs on one single GPU, which is not possible for transformer based models.

\begin{table}[h]
\centering
\scalebox{0.58}{
\begin{tabular}{lllll}
\toprule
               & \#Param & GPU Memory & Disk Space   & Source                                  \\\midrule
RoBERTa-Large  & 355M    & 2199MB     & 711MB      & fairseq (fp16)         \\
RoBERTa-Large  & 355M    & 2199MB     & 1.33GB     & HF transformers (fp32) \\
DistilBERT     & 66M     & 1123MB     & 256MB        & HF transformers                         \\
MobileBERT     & 25M     & 973MB      & 140MB        & HF transformers                         \\
DistilRoBERTa  & 85M     & 1181MB     & 316MB        & HF transformers                         \\
LSTM (2l-128d) & 53M     & 1051MB     & 212MB        & fairseq (fp32)                \\
CNN (1l-256d)  & 53M     & 1119MB     & 213MB        & fairseq (fp32)              \\
DAN            & 1001M   & 4655MB     & 3.99GB       & fairseq (fp32)        \\\bottomrule
\end{tabular}
}
\caption{Disk space and GPU memory required for each model.}\label{tab:disk}
\end{table}

\begin{table*}[t]
\vspace{0.0cm}
\centering
\scalebox{0.62}{
\begin{tabular}{l|rrr|crrr|crrr}
\toprule
& \multicolumn{3}{c|}{Parameter Count} & \multicolumn{4}{c|}{IMDB} & \multicolumn{4}{c}{SST-2} \\
                   & Total & Sparse & Dense & Acc. & GPU-fp32 & GPU-fp16 & CPU-fp32 & Acc. & GPU-fp32 & GPU-fp16 & CPU-fp32  \\\midrule
\ \ RoBERTa-Large      & 355M     & 51M             & 304M           & 96.3     & 28.9 (1x)   & 92.3 (1x)  & 1.4 (1x) & 96.2 & 267.3 (1x) & 610.2 (1x) & 22.2 (1x) \\
\ \ DistillBERT  & 66M & 23M & 43M & 92.2 & 175.8 (6x) & 334.7 (4x) & 10.7 (8x) & 90.8 & 828.5 (3x) & 1117.3 (2x) & 60.6 (3x) \\
\ \ MobileBERT  & 25M & 4M & 21M & 93.6 & 157.7 (5x) & 200.3 (2x) & 7.7 (6x) & 90.9 & 574.5 (2x) & 545.8 (1x) & 89.4 (4x) \\
$\star$DistillRoBERTa  & 83M & 39M & 44M & 95.9 & 176.4 (6x) & 569.8 (6x) & 7.8 (6x) & 94.2 & 636.5 (2x) & 771.7 (1x) & 185.9 (8x) \\
\midrule
$\star$LSTM (2l-512d) & 62M & 51M & 11M & 95.9 & 361.5 (12x) & 594.5 (6x)  & 30.6 (22x) & 93.9 & 4222.1 (14x) & 6281.4 (8x) & 394.3 (18x) \\
$\star$LSTM (2l-256d) & 56M & 51M & 5M  & 95.8 & 665.2 (23x) & 788.0 (9x) & 51.9 (37x) & 93.3 & 6361.5 (21x) & 7080.6 (9x) & 678.5 (31x) \\
$\star$LSTM (2l-64d) & 53M & 51M & 2M  & 94.0 & 818.5 (28x) & 808.5 (9x) & 101.4 (73x) & 92.8 & 7075.8 (24x) & 7384.1 (9x) & 1378.5 (62x)\\
$\star$LSTM (2l-4d) & 52M & 51M & <1M  & 93.1 & 812.9 (28x) & 817.0 (9x) & 146.4 (105x) & 88.3 & 7026.3 (24x) & 7521 (9x) & 2014.6 (91x)\\
$\star$CNN (1l-256d)  & 53M & 51M & 2M  & 89.2 & 3410.7 (109x) & 8427.1 (91x)  &  251.2 (181x) & 82.8 & 1323.5 (5x) & 1563.9 (3x) & 3820.4 (172x) \\ \midrule
$\star$DAN (ours)         & 1001M    & 1000M           & 1M             & 93.5     & 17557.9 (607x)  & 20888.1 (226x) & 922.6 (663x)  & 88.5 & 1745.5 (7x) & 1865.9 (3x) & 16478.6 (741x) \\\bottomrule
\end{tabular}
}
\caption{\textbf{Model Size and Inference Speed Comparison.} We report accuracy, inference speed (unit: samples per second) and relative speed compared to the teacher model (RoBERTa-Large). Our DAN model achieves competitive accuracy while achieving significant inference speed-up in various settings. $\star$ indicates the model is trained with task-specific distillation; no $\star$ indicates the model is trained with direct fine-tuning.
}
\label{table:speed_appendix}
\end{table*}

\begin{table*}[t]
\centering
\scalebox{0.7}{
\begin{tabular}{ll|cccccc}
\toprule
Notation & Description & IMDB & SST-2 & TREC & AGNews & CCom & WToxic\\\midrule
$\mathcal{V}_0$ & Top 1 million n-grams in $C$ and $D_{train}$ & \multicolumn{6}{c}{1,000,000} \\\midrule
$\mathcal{V}_1$ & All n-grams in $D_{train}$ & 10,109,522 & 262,417 & 89,358 & 7,156,063 & 116,143,462 & 15,805,923 \\
$\mathcal{V}_2$ & All n-grams in $D_{dev}$ & 9,843,369 & 39,666 & 5,995 & 662,665 & 8,987,055 & 6,958,457 \\\midrule

$\mathcal{V}_3$ & $\mathcal{V}_0\cap \mathcal{V}_1$ & 805,360 & 76,370 & 31,770 & 486,438 & 983,843 & 828,302 \\
\rowcolor{gray!20} & $|\mathcal{V}_0\cap \mathcal{V}_1|/|\mathcal{V}_0| \ (\%)$ & 80.54\% & 7.64\% & 3.18\% & 48.44\% & 98.38\% & 82.83\% \\
\rowcolor{gray!20} & $|\mathcal{V}_0\cap \mathcal{V}_1|/|\mathcal{V}_1| \ (\%)$ & 7.97\% & 29.10\% & 35.56\% & 6.80\% & 0.85\% & 5.24\% \\\midrule

$\mathcal{V}_4$ & $\mathcal{V}_0\cap \mathcal{V}_2$ & 792,251 & 15,395 & 3,461 & 123,247 & 740,286 & 671,985 \\
\rowcolor{gray!20} & $|\mathcal{V}_0\cap \mathcal{V}_2|/|\mathcal{V}_0| \ (\%)$ & 79.22\% & 1.54\% & 0.35\% & 12.32\% & 74.03\% & 67.20\% \\
\rowcolor{gray!20} & $|\mathcal{V}_0\cap \mathcal{V}_2|/|\mathcal{V}_2| \ (\%)$ & 8.05\% & 38.81\% & 57.73\% & 18.60\% & 8.23\% & 9.18\% \\\midrule

$\mathcal{V}_5$ & $\mathcal{V}_0\cap \mathcal{V}_1\cap \mathcal{V}_2$ & 690,790 & 8,804 & 1,840 & 113,311 & 739,920 & 638,833 \\
\rowcolor{gray!20} & $|\mathcal{V}_0\cap \mathcal{V}_1\cap \mathcal{V}_2|/|\mathcal{V}_2| \ (\%)$ & 7.01\% & 22.20\% & 30.69\% & 17.10\% & 8.23\% & 9.18\% \\\midrule

- & Average \# activated n-grams per instance & 496 & 16 & 17 & 68 & 103 & 144 \\
\bottomrule
\end{tabular}
}
\caption{Size of different sets of n-gram and their statistics of n-gram coverage.
}
\label{tab:ngram-stats}
\end{table*}


\section{Potential Risks}
It is risky to deploy DAN models to high-stakes applications (\textit{e.g.}, medical decisions) as the model lacks the ability of understanding long context (see case study in \S\ref{ssec:limitation}). DANs may raise fairness concerns: it lacks ability to understand the meaning of words in context, so it may learn spurious correlations such as overemphasis on group identifiers. We believe a thorough analysis is needed and bias mitigation methods such as \cite{bolukbasi2016man, kennedy2020contextualizing} are necessary for combating these issues.

\end{document}